%% file: main.tex
\documentclass[sigconf]{aamas} 
\usepackage{balance} % for balancing columns on the final page
\DeclareMathOperator*{\argmax}{arg\,max}

\usepackage{booktabs}
\usepackage{algorithm}
\usepackage[noend]{algpseudocode}
\usepackage{graphicx}
\usepackage{breqn}
\usepackage{subcaption}

\setcopyright{ifaamas}
\acmConference[AAMAS '23]{Proc.\@ of the 22nd International Conference
on Autonomous Agents and Multiagent Systems (AAMAS 2023)}{May 29 -- June 2, 2023}
{London, United Kingdom}{A.~Ricci, W.~Yeoh, N.~Agmon, B.~An (eds.)}
\copyrightyear{2023}
\acmYear{2023}
\acmDOI{}
\acmPrice{}
\acmISBN{}

\acmSubmissionID{947}

\title[AAMAS-2023 Formatting Instructions]{ExPoSe: Combining State-Based Exploration with Gradient-Based Online Search}

\author{Dixant Mittal}
\affiliation{
  \institution{National University of Singapore}
  \country{Singapore}}
\email{dixant@comp.nus.edu.sg}

\author{Siddharth Aravindan}
\affiliation{
  \institution{National University of Singapore}
  \country{Singapore}}
\email{siddharth.aravindan@comp.nus.edu.sg}

\author{Wee Sun Lee}
\affiliation{
  \institution{National University of Singapore}
  \country{Singapore}}
\email{leews@comp.nus.edu.sg}

\input{text/abstract.tex}

\newcommand{\BibTeX}{\rm B\kern-.05em{\sc i\kern-.025em b}\kern-.08em\TeX}

\begin{document}

%%% The following commands remove the headers in your paper. For final 
%%% papers, these will be inserted during the pagination process.

\pagestyle{fancy}
\fancyhead{}

%%% The next command prints the information defined in the preamble.

\maketitle 

%%%%%%%%%%%%%%%%%%%%%%%%%%%%%%%%%%%%%%%%%%%%%%%%%%%%%%%%%%%%%%%%%%%%%%%%
\input{text/introduction}
\input{text/background}
\input{text/method}
\input{text/experiments}
\input{text/related.tex}
\input{text/conclusion}
%%%%%%%%%%%%%%%%%%%%%%%%%%%%%%%%%%%%%%%%%%%%%%%%%%%%%%%%%%%%%%%%%%%%%%%%

%%% The next two lines define, first, the bibliography style to be 
%%% applied, and, second, the bibliography file to be used.

\bibliographystyle{ACM-Reference-Format} 
\bibliography{references}

%%%%%%%%%%%%%%%%%%%%%%%%%%%%%%%%%%%%%%%%%%%%%%%%%%%%%%%%%%%%%%%%%%%%%%%%

% \clearpage
% \onecolumn
% \appendix
% \input{appendix/policy-gradient-derivation}
% \input{appendix/domains}
% \input{appendix/networks}
% \input{appendix/visualisation}
% \input{appendix/atari-scores}

\end{document}

%% file: text/abstract.tex
\begin{abstract}
Online tree-based search algorithms iteratively simulate trajectories and update action-values for a set of states stored in a tree structure. It works reasonably well in practice but fails to effectively utilise the information gathered from similar states. Depending upon the smoothness of the action-value function, one approach to overcoming this issue is through online learning, where information is interpolated among similar states; Policy Gradient Search provides a practical algorithm to achieve this. However, Policy Gradient Search lacks an explicit exploration mechanism, which is a key feature of tree-based online search algorithms. In this paper, we propose an efficient and effective online search algorithm called Exploratory Policy Gradient Search (ExPoSe), which leverages information sharing among states by updating the search policy parameters directly, while incorporating a well-defined exploration mechanism during the online search process. We evaluate ExPoSe on a range of decision-making problems, including Atari games, Sokoban, and Hamiltonian cycle search in sparse graphs. The results demonstrate that ExPoSe consistently outperforms other popular online search algorithms across all domains. The ExPoSe source code is available at \textit{\url{https://github.com/dixantmittal/ExPoSe}}.
\end{abstract}
\keywords{Exploration; Online Tree Search; Reinforcement Learning}

%% file: text/introduction.tex
\section{Introduction} \label{section:introduction}
Online search algorithms have shown good performance on complex search problems such as Chess \cite{silver2017mastering} and Go \cite{silver2016mastering}. Monte Carlo Tree Search (MCTS) is a popular online tree search algorithm that builds a tree structure to maintain a Monte Carlo average of Q-values by simulating different scenarios from the input state. Some variants of MCTS, like Upper Confidence bound applied to Trees (UCT) \cite{kocsis2006bandit}, use a best-first search approach to explore regions of the search tree that could potentially return better rewards. UCT achieves this by adding an exploration bonus to the Q-values of the state-action nodes based on their visitation counts while selecting an action during the simulation phase. This encourages exploration by occasionally prompting it to choose a less-visited child node. Predictor-UCT (PUCT) \cite{rosin2011multi} is a variant of UCT used in AlphaGo \cite{silver2016mastering} that incorporates prior domain knowledge by scaling the exploration bonus of an action with its prior probability, which is predicted by a learnt policy network. UCT-type tree search methods have a well-defined exploration-exploitation trade-off mechanism that is easy to implement and perform well in practice. 

Unfortunately, tree search methods have a significant limitation: they do not fully exploit the information gained from each simulation. Specifically, the information obtained from a rollout trajectory is only used to update the action-value estimates of the nodes in the path traversed by the agent, and the rest of the nodes in the tree are ignored. As a result, the search algorithm cannot share the knowledge gained from the expanded leaf nodes to similar states. This problem becomes more pronounced when the search budget is limited, severely limiting the performance of tree search methods.

Online learning methods provide a way to distribute search information obtained from a trajectory to the entire state space. They directly update the parameters of the function being learned, and the information gained during each update is reflected across the state space due to the generalization capabilities of the learned function approximator. Online gradient descent is a common online learning method, and policy gradient can be used to produce the online gradient from a trajectory. A search algorithm can be developed by iteratively simulating a trajectory from the input state using the current policy and applying policy gradient updates to improve the policy after each trajectory. Policy Gradient Search (PGS)~\cite{anthony2019policy} is one such example and performs well for many combinatorial problems \cite{anthony2019policy,bello2016neural}. However, a key issue with PGS is the inherent lack of a proper exploration mechanism.
 
We propose a novel online search algorithm called \textbf{Ex}ploratory \textbf{Po}licy Gradient \textbf{Se}arch (ExPoSe) that leverages the benefits of both tree search and online learning methods. ExPoSe utilises a tree structure to store state statistics, such as visitation counts and value estimates, allowing it to define a robust exploration mechanism and exploit the valuable information from the subtree at each node in the search tree. Moreover, it incorporates online learning to efficiently distribute the information obtained from a new trajectory across the entire state space.

ExPoSe iteratively simulates trajectories from the input state using a simulation policy, and maintains a visitation count for each encountered state. To achieve exploration, the simulation policy modifies the learned policy using a state-based exploration term that is inversely proportional to the visitation count of the state. The exploration term is added to the unnormalised log probabilities of actions predicted by the learned policy network, thereby enabling a bonus-based exploration mechanism. Moreover, ExPoSe propagates the information obtained from the simulated trajectory by directly updating the learnt policy parameters using policy gradient methods. 

After conducting experiments, we observe that ExPoSe outperforms PGS when exploration is incorporated into the policy gradient search. However, there are still two issues that need to be addressed to further improve the agent's performance. The first issue is related to the fact that the trajectory is generated using a simulation policy, which differs from the target policy being learned due to the addition of an exploration term. This difference can lead to incorrect gradient computation if we use policy gradient methods, such as REINFORCE~\cite{williams1992simple}. To address this issue, we use importance sampling to correct the off-policy evaluation and obtain an unbiased estimate of the gradients. The second issue is that policy gradient methods, like REINFORCE, suffer from high variance due to the stochastic nature of the policy. To reduce the variance of the policy gradients while being unbiased, we use a baseline term. The value estimate of each state in the trajectory can serve as an effective baseline and can be used to construct an advantage while computing the policy gradients. This technique generally works well in practice. Fortunately, the tree structure used in ExPoSe enables us to efficiently compute the value estimates of a state from its subtree by using the Bellman equation.

In our empirical study, we investigate the efficacy of incorporating these key ideas from tree-based and gradient-based search methods on a wide range of decision-making problems, including popular goal-based problems like Sokoban, Hamiltonian cycle search in sparse graphs, and grid navigation, as well as image-based games in the Atari 2600 test suite. We find that ExPoSe consistently outperforms existing online search algorithms on all the domains, demonstrating its ability to leverage the strengths of tree-based and gradient-based search methods to enhance the agent's performance. These results indicate the potential of ExPoSe as a general-purpose online search algorithm for a variety of complex decision-making problems.

%% file: text/background.tex
\section{Background} \label{section:background}
\subsection{Markov Decision Process}
Many real-world tasks can be formulated as Markov decision processes (MDPs), where the agent interacts with the environment by choosing a sequence of actions to reach the goal. In MDPs, the agent observes a state described by a collection of variables that provide relevant details for action selection. Given a state, the agent chooses an action determined by a policy $\pi$, which is a function that maps every state $s$ in the state space to an action $a$ and receives a feedback reward $r$ from the environment. A value function $V^\pi(s)$ represents the expected sum of all rewards the agent accrues by following a policy $\pi$ when starting from state $s$. Similarly, a Q-function (or action-value function) $Q^\pi(s, a)$ represents the expected sum of rewards obtained when the agent chooses the action $a$ in the state $s$ but follows the policy $\pi$ in the subsequent states until the episode terminates. To solve the MDP, the agent has to find a policy $\pi^*$, which maximises the value function $V^\pi(s)$ for all states $s$,  i.e., 
\begin{equation*}
    \pi^* = \argmax_\pi V^\pi(s)
\end{equation*}
In general, finding the optimal policy for all the states is a challenging problem in an MDP with a large state space. Therefore, agents solving such MDPs are often assisted by online search methods, such as UCT \cite{kocsis2006bandit} and branch-and-bound \cite{lawler1966branch} search. These online search methods are designed to help the agent determine an approximately optimal action for a given state.

\subsection{Monte Carlo Tree Search} \label{subsection:mcts}
Monte Carlo Tree Search (MCTS) is an online tree-based search algorithm that improves the policy by iteratively simulating trajectories and expanding the search tree towards a more promising search space. Each tree node represents a state, and the root node represents the input state for which the agent has to output an action. The children of any node in the tree represent the states that are reachable from that node by taking a single action represented by the corresponding branch. In every iteration, MCTS performs the following operations in order: \textit{selection}, \textit{initialisation} and \textit{backup}. This process is repeated until termination, determined by either a fixed number of search iterations or a maximum search time allocated for each step. At the end of the search process, the agent chooses the action corresponding to the highest action-value at the root node.

The selection phase in MCTS is flexible and has been modified to create different variants, such as UCT~\cite{kocsis2006bandit} or PUCT \cite{rosin2011multi}. These variants usually differ in their exploration mechanism and have achieved good performances in different problems.

\subsubsection{Selection} \label{subsubsection:selection} \hfill \\
During the selection phase, the agent uses a tree simulation policy to traverse a path from the root node to a leaf node in the search tree. The tree simulation policy uses various node statistics for a given node, such as a Monte-Carlo average of Q-values or visitation counts, to select an action. Tree simulation policies may also include a bonus term favouring actions that are likely to have the highest Q-value but are selected less frequently. UCT, for example, uses the UCB1~\cite{kocsis2006bandit} formulation to add an exploration bonus $U(s,a)$ to the estimated Q-values $Q(s,a)$. More precisely, $U(s,a) \propto \sqrt{\dfrac{\log N(s)}{N(s,a)}}$, where $N(s)$ is the number of times the agent has visited the node s, while $N(s,a)$ is the number of times the agent has chosen the action $a$ at node $s$. On the other hand, PUCT uses a policy $\pi_{\theta}$, which is learnt offline, to scale the exploration bonus added to the Q-value corresponding to action branch $a$ by  $\pi_{\theta}(a|s)$, i.e. $U(s, a) \propto \dfrac{\pi_{\theta}(a|s)}{1 + N(s, a)}$. The agent uses the tree simulation policy until it reaches a leaf node.

\subsubsection{Expansion} \label{subsubsection:expansion} \hfill \\
The leaf node encountered at the end of the simulation phase is expanded by adding its children to the search tree. These new nodes are initialised with a prior value estimate given by an offline learnt value function approximator. After adding these nodes to the tree, a rollout trajectory is played using a rollout policy from the leaf node to get an unbiased estimate of its value. Usually, the rollout policy is trained offline on a large data set to get a better estimate of the node's value. However, some variants, like AlphaGo\cite{silver2016mastering}, may use a faster but less accurate policy to perform the rollout.

\subsubsection{Backup} \label{subsubsection:backup} \hfill \\
After completing the rollout from the leaf node, the rollout value is propagated upwards to update the Q-value estimates along the path to the root. The common variants of MCTS maintain a Monte-Carlo average of the Q-values for each node and can trivially incorporate the new rollout value into the average. Some variants may also maintain an empirical variance of the observed rollout values, which could be used to compute the upper confidence bound during the simulation phase.

\subsection{Policy Gradient Search} \label{subsection:pgs}
Policy Gradient Search (PGS) is an online gradient-based search algorithm that iteratively improves a simulation policy, parameterised by $\theta$, using policy gradient methods. Each iteration starts from the root and uses a simulation policy to simulate a trajectory. We define an objective function $J_\theta(\tau)$ that measures the policy's performance based on the rewards obtained from the trajectory $\tau$ i.e.:
\begin{equation*}
    J_\theta(\tau) = \mathbb{E}_{\tau \sim \pi_\theta(\tau)} r (\tau)
\end{equation*}
where $r (\tau)$ is the sum of rewards obtained for the trajectory $\tau$. Given this objective function, we can compute the gradient of the policy parameters using policy gradient methods, such as REINFORCE \cite{williams1992simple}, and update the policy parameters using an optimiser such as stochastic gradient descent. The parameter update is given by:
\begin{equation*}
    \theta \leftarrow \theta + \alpha \ \dfrac{1}{N} \sum_{i=1}^N \Bigg[ \sum_{t = 1}^T \nabla_\theta \log \pi_\theta (a_{i,t} | s_{i,t}) \sum_{t'= 1}^T r(s_{i,t'},a_{i,t'})  \Bigg]
    % \Bigg[ \sum_{t=1}^T \nabla_\theta \log \pi_\theta (a_t | s_t) \sum_{t'= 1}^T r(s_{t'},a_{t'}) \Bigg]
\end{equation*}
where 
\begin{itemize}
    \item $\pi_\theta$: parameterised policy with parameters $\theta$,
    \item $\sum_{t'= 1}^T r(s_{i,t'},a_{i,t'})$: sum of rewards for trajectory $\tau_i$.
    \item $s_{i,t}$ and $a_{i,t}$ are the state and the action respectively at timestep $t$ in trajectory $\tau_i$.
\end{itemize}

Policy Gradient Search maintains the values observed from simulations at the root node and uses PUCT-type formulation to select the first action $a_{root}$ at root node $s_{root}$ i.e.
\begin{equation*}
    a_{root} = \argmax_a \Bigg[ Q(s_{root}, a) + c \pi(a|s_{root}) \dfrac{\sqrt{N(s_{root})}}{1 + N(s_{root},a)}  \Bigg]
\end{equation*}
However, the subsequent actions are sampled from the simulation policy only.

Neural Networks are commonly used to approximate policy in gradient-based search. However, due to the high computational cost of backpropagation, PGS calculates gradients for only the policy head parameters. Additionally, Anthony et al.~\cite{anthony2019policy} recommend freezing earlier layers to reduce the FLOPS required for the backward pass, resulting in the parameter optimisation step being computationally insignificant compared to other search operations.

%% file: text/method.tex
\section{Exploratory Policy Gradient Search} \label{section:method}
Both tree-based and gradient-based online search algorithms improve the agent's performance when combined with offline learnt prior policy and value functions. We identify and empirically verify that the key properties that improve the performance of these methods are complementary to each other. These properties are:
\begin{enumerate}
    \item Interpolation of the updated search information across the state space helps in adapting the search policy for similar states.
    \item A well-defined exploration mechanism that can efficiently balance the exploration-exploitation trade-off is important for the online search.
\end{enumerate}

Along the same lines, we propose an efficient and effective search method named \textbf{Ex}ploratory \textbf{Po}licy Gradient \textbf{Se}arch (ExPoSe). ExPoSe iteratively simulates trajectories from the input state using a simulation policy and directly updates the parameters of the search policy using policy gradient methods, effectively interpolating the information obtained from the simulated trajectories across the state space. Moreover, it maintains a tree structure that stores state statistics, like visitation counts and the value estimate of a state, of all the states encountered during the simulation. ExPoSe uses a visitation counts-based exploration incentive that encourages the simulation policy to discover new states during the search. Additionally, the tree structure allows a better estimation of the state values using the Bellman equation, which can be used to reduce the variance in policy gradient methods. We discuss ExPoSe in detail in the following sections.

\subsection{Policy Gradients} \label{subsection:policy-gradient}
The overall structure of ExPoSe is similar to PGS (Section \ref{subsection:pgs}). ExPoSe iteratively simulates trajectories from the input state using a simulation policy, $\phi_{sim}(s,a)$, which is a combination of a parameterised policy network $\phi_\theta(s,a)$ and an exploration term (defined in Section \ref{subsection:state-based-exploration}). After each simulation, the information obtained from the trajectory $\tau$ is used to optimise the objective function $J_\theta(\tau)$ by computing gradients for the parameters $\theta$ of the policy network $\phi_\theta(s, a)$ using REINFORCE i.e.
\begin{dmath}\label{eq:reinforce-vanilla}
    \nabla_\theta J(\tau) 
    = \mathbb{E}_{\tau \sim \pi_\theta(\tau)} \Bigg[ \sum_{t = 1}^T \nabla_\theta \log \pi_\theta (a_{t} | s_{t}) \sum_{t'= 1}^T r(s_{t'},a_{t'}) \Bigg]
    % = \dfrac{1}{N} \sum_i^N \Bigg[ \sum_{t = 0}^T \nabla_\theta \log \pi_\theta (a_{i,t} | s_{i,t}) \sum_{t'= 1}^T r(s_{i,t'},a_{i,t'}) \Bigg]
\end{dmath} 
where 
\begin{itemize}
    \item $J_\theta(\tau)$: objective function to compute policy gradients.
    \item $\tau$: sequence of state action pairs i.e. $\tau = (s_1,a_1,...,s_T,a_T)$.
    \item $r(s_{t'},a_{t'})$: rewards obtained at timestep $t$.
    \item $\pi_\theta$: normalised probability i.e. $\pi_\theta(a|s) = \dfrac{\exp(\phi_\theta(s,a))}{\sum_a \exp(\phi_\theta(s,a))}$
\end{itemize}
Since we have a limited search budget, we set $N=1$ and use a single trajectory $\tau$ to approximate the expectation. Hence, we can simplify the equation
\begin{equation} \label{eq:reinforce-simplified}
    \nabla_\theta J(\tau) = \Bigg[ \sum_{t = 1}^T \nabla_\theta \log \pi_\theta (a_{t} | s_{t}) \sum_{t'= 1}^T r(s_{t'},a_{t'}) \Bigg]
\end{equation}
Correspondingly, the policy network parameters $\theta$ can be updated using the gradient ascent algorithm.
\begin{equation} \label{eq:vanilla-pg-update}
    \theta \leftarrow \theta + \alpha \Bigg[ \sum_{t = 1}^T \nabla_\theta \log \pi_\theta (a_{t} | s_{t}) \sum_{t'= 1}^T r(s_{t'},a_{t'}) \Bigg]
\end{equation}
where $\alpha$ is the tunable learning rate.

We repeat this process of iterating through simulation and policy optimisation until we exhaust the search budget. Since we optimise the policy parameters directly using REINFORCE, the updated information is reflected on the policy determined by $\phi_{\theta}(s, a)$ for every state $s$ in the state space. 

\subsection{State-based Exploration} \label{subsection:state-based-exploration}
Policy Gradient Search's major drawback is the inherent lack of a well-defined exploration mechanism. Larma et.al.~\cite{larma2021improving} suggests a simple way to induce exploration in the policy by adding entropy regularisation in equation~\ref{eq:vanilla-pg-update}. However, entropy regularisation is state-agnostic and induces exploration by increasing the randomness in the policy throughout the state space, which may not be desirable. We suggest and empirically verify that a better way to balance the exploration-exploitation trade-off is to use a principled approach based on state statistics like visitation counts. We take inspiration from UCB1~\cite{kocsis2006bandit} and propose to augment the simulation policy used in ExPoSe with a state-based exploration term $E(s,a)$, for a state $s$ and corresponding action $a$, which is inversely proportional to the visitation count $N(s,a)$, i.e. $E(s,a) = \dfrac{c}{1 + N(s,a)}$, where $c$ is a tunable hyperparameter. ExPoSe maintains the visitation counts $N(s, a)$ of all the state-action pairs encountered during the simulations. The simulation policy at any state is derived by combining the logits $\phi_{\theta}(s, a)$ predicted by the parameterised policy network and the exploration term $E(s, a)$ as follows: 
\begin{dmath*}
\phi_{sim}(s,a) = \phi_{\theta}(s,a) + E(s,a) \\
                = \phi_{\theta}(s,a) + \dfrac{c}{1 + N(s,a)}
\end{dmath*}
We apply the softmax operation over the logits $\phi_{sim}(s, a)$ to get the normalised simulation policy distribution $\pi_{sim}(s,a)$ i.e.
\begin{equation*}
    \pi_{sim}(a|s) = \dfrac{\exp(\phi_{sim}(s,a))}{\sum_a \exp(\phi_{sim}(s,a))}
\end{equation*}

We sample actions from this normalised simulation policy distribution to simulate a trajectory. The exploration term encourages less frequently selected actions to be sampled with higher probability, which leads to the exploration of new state space. Furthermore, the exploration term for a certain state-action pair decreases as its visitation count increases, allowing for an efficient transfer from exploration to exploitation.

\subsection{Importance Sampling} \label{subsection:importance-sampling}
If the trajectory $\tau$ is generated using the exploration-induced simulation policy $\pi_{sim}(a|s)$, the policy gradient computed in equation~\ref{eq:reinforce-simplified} is biased because it is being computed off-policy, i.e. the policy used to generate the trajectory is different from the parameterised policy used to compute policy gradients. However, we can fix this issue and compute an asymptotically unbiased estimate of the gradients using importance sampling. Let us assume that the simulation policy is the behaviour policy, i.e. the policy that generates the data, and the policy network is the target policy, i.e. the policy which will be optimised. We can compute the policy gradients for the target policy by reweighing the objective function, $J_\theta(\tau)$, with the likelihood of the trajectory under the behaviour policy and the target policy. The importance sampling ratio $w$ is computed as:
\begin{equation*}
    w = \dfrac{\text{P}_{\theta}(\tau)}{\text{P}_{sim}(\tau)} = \prod_{t=1}^T \frac{\pi_\theta(a_t | s_t)}{\pi_{sim}(a_t | s_t)}
\end{equation*}

We can re-write equation~\ref{eq:reinforce-simplified} along with the importance weights as follows:
\begin{equation} \label{eq:objective-with-IS}
    \nabla_\theta J(\tau) = \Bigg[ \sum_{t = 1}^T \prod_{j=1}^T \frac{\pi_\theta(a_j|s_j)}{\pi_{sim}(a_j|s_j)}   \nabla_\theta \log \pi_\theta (a_{t} | s_{t}) \sum_{t'= 1}^T r(s_{t'},a_{t'}) \Bigg]
\end{equation}

We empirically verify that importance sampling helps in improving the agent's performance.

\subsection{Tree-based Value Approximation} \label{subsection:tree-based-value}
We use Monte-Carlo sampling to compute the total reward used in REINFORCE's objective function in equation \ref{eq:reinforce-vanilla}. However, even if we start from the same state, the sampled trajectories can lead to different rewards due to stochasticity in the policy. Consequently, the variance in estimating the total reward is high and can adversely affect policy optimisation.

Previous works\cite{mnih2016asynchronous,williams1992simple,levine2013guided,sutton2018reinforcement} have used causality and the Markov property of these problems to discover tricks to reduce the variance of the policy gradient while keeping the estimation unbiased. Firstly, we can replace the total rewards in equation \ref{eq:objective-with-IS} with the sum of rewards obtained from state at timestep $t$ onwards as the action at timestep $t$ cannot affect the rewards obtained prior to timestep $t$. Similarly, the importance sampling ratio at timestep $t$ is not affected by future actions. Hence, we can rewrite equation \ref{eq:objective-with-IS} as follows:
\begin{dmath} \label{eq:objective-with-causality}
    \nabla_\theta J(\tau) = \Bigg[ \sum_{t = 1}^T \prod_{j=1}^t \frac{\pi_\theta(a_j|s_j)}{\pi_{sim}(a_j|s_j)}   \nabla_\theta \log \pi_\theta (a_{t} | s_{t}) \sum_{t'= t}^T r(s_{t'},a_{t'}) \Bigg] \\
    = \Bigg[ \sum_{t = 1}^T w_t   \nabla_\theta \log \pi_\theta (a_{t} | s_{t}) \ \hat{Q}_t \Bigg]
\end{dmath}
where $w_t = \prod_{j=1}^t \frac{\pi_\theta(a_j|s_j)}{\pi_{sim}(a_j|s_j)}$ and $\hat{Q}_t = \sum_{t'= t}^T r(s_{t'},a_{t'})$ be the observed Q-value at timestep $t$.

Secondly, we can subtract a baseline term from the observed Q-values~\cite{mnih2016asynchronous}. A popular baseline is the value predictions of the states observed in a trajectory. If we subtract the value of a state from the observed Q-value, we get the advantage value of taking an action over its expected value. The advantage value has lower variance while being unbiased and works well in practice~\cite{mnih2016asynchronous}. We can rewrite the equation \ref{eq:objective-with-causality} as follows:
\begin{equation} \label{eq:final-objective}
    \nabla_\theta J_\theta(\tau) = \Bigg[ \sum_{t = 1}^T w_t   \nabla_\theta \log \pi_\theta (a_{t} | s_{t}) \Big(\hat{Q}_t - V(s_t) \Big)\Bigg]
\end{equation}

On the other hand, a tree structure can store state statistics that allow us to efficiently maintain an estimate of the values of the observed states by backing up the observed value from a trajectory. The tree structure also enables us to recursively apply the Bellman equation on all the states in the trajectory as follows:
\begin{equation*}
    V_{tree}(s) = \max_a \big[r(s,a) + \gamma \sum_{s'} P(s'|s,a) V_{tree}(s') \big]
\end{equation*}
where 
\begin{itemize}
    \item $s'$ is the next state observed after taking action $a$ in state $s$.
    \item $V(s)$ is the value of state $s$.
    \item $r(s,a)$ is the reward obtained after taking action $a$ in state $s$.
\end{itemize}

Finally, we can write the combined policy gradient update as follows:
\begin{equation} \label{eq:final-policy-update}
    \theta \leftarrow \theta + \alpha \Bigg[ \sum_{t = 1}^T w_t   \nabla_\theta \log \pi_\theta (a_{t} | s_{t}) \Big(\hat{Q}_t - V_{tree}(s_t) \Big)\Bigg]
\end{equation}

In policy gradient-based search methods, it is a common practice to freeze all the parameters except the parameters of the last network layer to speed up the search and avoid slowdowns due to excessive gradient computation \cite{anthony2019policy}. We follow the same process in ExPoSe and calculate the gradients for the parameters of the last network layer only.

%% file: text/experiments.tex
\input{tables/sokoban} 
\input{tables/hamiltonian}
\section{Experiments} \label{section:experiments}
\subsection{Experiment Setup}
For our experiments, we set a maximum number of search iterations per step for each method to limit the search budget. All the online search methods use the same policy and value function approximators for each problem to enable comparison of policy improvement due to the strengths of each search method. We measure the success rate to evaluate agent performance in goal-based planning problems, i.e., the fraction of instances where the agent reaches the goal state. In contrast, we use Human Normalised Score (HNS) and Baseline Normalised Score (BNS) to evaluate agent performance on Atari games.

Each online search method has its set of adjustable hyperparameters. We use $10\%$ of the testing data as the validation set and the remaining $90\%$ as the holdout test set to report the final evaluation score. We select the best set of hyperparameters using grid search on a log-linear scale with the validation set. Finally, we use the best set of hyperparameters for each method to evaluate the agent's performance on the holdout test set.

\subsection{Baselines}
We use popular tree-based and gradient-based online search algorithms, PUCT and PGS, as primary baselines for comparison. Below are the specific implementation details for each method:
\begin{itemize}
    \item \textbf{PUCT}: We implement the standard version of PUCT as described in Section \ref{subsection:mcts}.
    \item \textbf{PGS}: We implement the standard version of PGS as described in Section \ref{subsection:pgs}. Additionally, we add entropy and L2 regularization to the policy to prevent it from converging to a deterministic policy. Entropy regularization also induces exploration by increasing the randomness in the policy \cite{larma2021improving}.
    \item \textbf{ExPoSe}: We implement ExPoSe as described in Section \ref{section:method}. We also add entropy and L2 regularisation as described for PGS above.
\end{itemize}

\subsection{Domains}
We evaluate ExPoSe and other online search methods on two types of decision-making problems:
\begin{itemize}
    \item A set of goal-based planning problems that includes Sokoban, Hamiltonian cycle search in sparse graphs and 2D grid navigation.
    \item Atari 2600 benchmarking suite, which is a set of image-based games.
\end{itemize}

We describe each domain in brief as follows:
\subsubsection{Sokoban} \label{subsubsection:sokoban} \hfill \\
Sokoban is a classic puzzle game where an agent must move boxes to designated goal positions without hitting the walls. Because actions have irreversible consequences, Sokoban poses a complex planning problem. Sokoban has been widely used for experimentation in recent research \cite{guez2018learning,guez2019investigation}.

To ensure standardized comparison across methods, we use the Boxoban dataset \cite{boxobanlevels} as described in \cite{guez2019investigation}. The dataset consists of Sokoban instances from three difficulty levels: hard, medium, and unfiltered. Rather than using a reinforcement learning algorithm, such as A3C \cite{mnih2016asynchronous}, to train both the policy and value function approximators (as done in \cite{guez2019investigation}), we use Expert Iteration (ExIt) \cite{anthony2017thinking} to train the policy and value function due to its more stable training regime. The ExIt algorithm iteratively executes two processes: (1) using an expert policy to collect better data and (2) using the collected data to train the policy and value function. We use PUCT with a high number of search iterations combined with a partially trained policy and value function as the expert policy. We use training samples from the unfiltered and medium problem sets to collect expert data. Rather than starting from random weights, we pre-train the policy and value function approximators on data collected using $\text{A}^*$ search. To compare different search methods, we focus on the Boxoban hard test set, as almost all test instances in the unfiltered and medium test sets are solvable by all search methods.

\subsubsection{Hamiltonian Cycle} \label{subsubsection:hamiltonian-cycle} \hfill \\
A Hamiltonian cycle is a path in a graph that visits each node exactly once and returns to its starting node. Finding a Hamiltonian cycle in a sparse graph is a challenging planning problem because choosing the wrong action at the beginning can have long-term consequences, leading to the algorithm getting stuck with no node to visit next. Therefore, it is crucial to anticipate the future effects of actions and create a plan that can ultimately complete a Hamiltonian cycle.

To generate expert data for this problem, we randomly permute the nodes, connect them to create a cycle, and add random edges to create a sparse graph with a known Hamiltonian cycle. We create a training dataset of $10,000$ graphs with $50$ nodes and $50\%$ sparsity. To make the problem harder for the test set, we generate graphs with 50 nodes and $10\%$ sparsity.

\subsubsection{Grid Navigation} \label{subsection:navigation} \hfill \\
Grid navigation is a task that requires an agent to navigate a 2D grid to reach a goal position while avoiding obstacles. The agent has access to information about its current and goal positions, as well as an environment map that indicates the locations of obstacles. The main challenge is to find a path from the current position to the goal that may require taking detours due to obstacles blocking the direct path.

We conduct experiments on a 2D grid of size $50 \times 50$. To create a training dataset, we generate $100,000$ random maps where obstacles occupy each cell with a probability of $0.25$. During map generation, we randomly select the agent's starting and goal positions using rejection sampling such that the shortest distance between the starting and goal positions is more than $100$ units, and their Manhattan distance is more than $50$ units. This ensures that the generated levels include a detour from the direct path. Additionally, we create a test set of $5,000$ instances with even more challenging and denser maps by increasing the probability of an obstacle occupying a cell to $0.50$. The shortest path to the goal in these instances involves significant detours. Since the state space is small, any shortest path algorithm can act as an expert policy to generate the training data. We use this toy problem primarily to visualise the state space explored by different search methods (refer to Figure \ref{figure:exploration} in the appendix).

\input{tables/navigation}
\input{tables/atari}
\subsubsection{Atari 2600 Benchmarking Suite} \label{subsection:atari} \hfill \\
Atari 2600 is a suite of image-based games commonly used to evaluate agents' policies \cite{Mnih2013PlayingAW,mnih2016asynchronous}. In this suite, the agent receives an image as an observation from the simulator, and the goal is to score as high as possible before the episode ends. To represent the state, popular methods stack a sequence of images.

We use a set of 50 Atari 2600 games (as mentioned in \cite{badia2020agent57}) to evaluate the online search methods. To do this, we use a prior policy and value functions for each game, trained through A3C \cite{mnih2016asynchronous}, and provided by PFRL \cite{fujita2021pfrl}. For evaluation, we apply Atari wrappers and agents from PFRL, and each online search method uses the same prior policy and value networks. To measure the agents' performance, we use Human Normalised Score (HNS) (Eq. \ref{eq:hns}) and Baseline Normalised Score (BNS) (Eq. \ref{eq:bns}), averaged over all the games.
\begin{equation}\label{eq:hns}
    HNS = \dfrac{S_\pi - S_R}{S_H - S_R}
\end{equation}

\begin{equation}\label{eq:bns}
    BNS = \dfrac{S_\pi - S_R}{S_B - S_R}
\end{equation}
where $S_\pi, S_R, S_H, S_B$ represents the score achieved by the agent when following the online search policy $\pi$, a random policy, the human expert policy and the baseline policy, respectively.

We limit each search method to a maximum of $10$ iterations during each prediction step. To speed up evaluation, we simulate a trajectory of $20$ steps and use the value predicted by the value network at the last step as a bootstrap. In Table \ref{table:atari}, we report both the mean and median scores achieved by the agent. The appendix provides a detailed list of scores achieved by the agent on each of the 50 games in the Atari test suite.

\subsection{Results}
We present our experimental results in Table \ref{table:sokoban}, Table \ref{table:hamiltonian-cycle}, Table \ref{table:navigation}, and Table \ref{table:atari} for Sokoban, Hamiltonian cycle search, grid navigation, and Atari 2600 games respectively. We find that ExPoSe consistently outperforms all baselines across a diverse set of testing domains. We also observe that all online search methods improve the performance of the prior policy in goal-based problems, and ExPoSe outperforms PUCT and PGS given the same search budget. Moreover, ExPoSe exhibits a higher margin of improvement over baselines when we allow a small search budget. Our results also indicate that ExPoSe performs better than PGS in 38 out of 50 Atari games, with better scores on HNS and BNS averaged over all the games (see Table \ref{table:atari}).

\subsubsection{Does information sharing among states help in online search?} \hfill \\
Methods that directly update the information into the parameters of the policy, i.e. PGS and ExPoSe, outperform PUCT, which does not share information across the search tree. Thus, we can empirically conclude that information sharing across the states helps improve the policy obtained using an online search.

\subsubsection{Does state-based exploration help improve the performance of gradient-based online search?} \hfill \\
ExPoSe, which combines state-based exploration with gradient-based online search, outperforms PGS with entropy regularization across all domains. This further supports the importance of a well-defined state-based exploration mechanism in gradient-based online search methods.

\subsection{Ablation Studies} \label{subsection:ablation-study}
We conduct an ablation study on Sokoban to evaluate the contribution of the improvements described in Section \ref{subsection:importance-sampling} and \ref{subsection:tree-based-value}. To perform the comparison, we modify the original ExPoSe implementation and create the following implementations:
\begin{itemize}
    \item \textbf{ExPoSe without Importance Sampling}: We remove the importance sampling ratio term while computing the policy gradients. The corresponding parameter update is given by:
    \begin{equation*}
        \theta \leftarrow \theta + \alpha \Bigg[ \sum_{t = 1}^T \nabla_\theta \log \pi_\theta (a_{t} | s_{t}) \Big(\hat{Q}_t - V_{tree}(s_t) \Big)\Bigg]
    \end{equation*}
    
    \item \textbf{ExPoSe without Value Baseline}: We exclude any baseline term while computing the policy gradients. The corresponding parameter update is given by:
    \begin{equation*}
        \theta \leftarrow \theta + \alpha \Bigg[ \sum_{t = 1}^T w_t\nabla_\theta \log \pi_\theta (a_{t} | s_{t}) \ \hat{Q}_t \Bigg]
    \end{equation*}
    
    \item \textbf{ExPoSe with Value Network Baseline}: We replace the tree-based value estimation term with an offline learnt value function as the baseline while computing the policy gradients. The corresponding parameter update is then given by:
    \begin{equation*}
        \theta \leftarrow \theta + \alpha \Bigg[ \sum_{t = 1}^T w_t \nabla_\theta \log \pi_\theta (a_{t} | s_{t}) \Big(\hat{Q}_t - V_{net}(s_t) \Big)\Bigg]
    \end{equation*}
\end{itemize}
We present the results of our ablation study in Table \ref{table:ablation}, which allows us to address the following questions:
\subsubsection{Does importance sampling help improve the computation of the policy gradients?} \hfill \\
In Table \ref{table:ablation}, we compare the performance of the original \textit{ExPoSe} with that of \textit{ExPoSe without importance sampling}. The results demonstrate that the original \textit{ExPoSe} consistently outperforms the \textit{ExPoSe without importance sampling}, providing empirical evidence for the effectiveness of importance sampling in computing policy gradients.

\subsubsection{Does using state values as a baseline help reduce the variance and improve the agent's performance?} \hfill \\
Table \ref{table:ablation} shows that the \textit{ExPoSe} variant without a value baseline consistently performs worse than the original \textit{ExPoSe} and the modified version with a value network baseline. These results empirically demonstrate that incorporating a value baseline term in policy gradient computation helps reduce variance and improve agent performance.

\subsubsection{Does tree-based value estimation perform better than an offline learnt value network?} \hfill \\
Table \ref{table:ablation} shows that the \textit{ExPoSe} algorithm with tree-based value estimation consistently outperforms the \textit{ExPoSe with value network baseline} by a small margin. These results suggest that incorporating a tree-based value estimation mechanism can help improve the agent's performance when compared to using an offline-learned value network as a baseline.
\input{tables/ablation}

%% file: tables/sokoban.tex
\begin{table*}
    \caption{Comparison of test performance on Sokoban (Boxoban hard set) measured in success rate (i.e. \% of test instances solved) for different search methods ["\#iterations" stands for number of search iterations]}
    \label{table:sokoban}
    \centering
    \begin{tabular}{c@{\hspace{8\tabcolsep}}c@{\hspace{10\tabcolsep}}c@{\hspace{10\tabcolsep}}c}
    \toprule
    Search Method                       &\#iterations=10           &\#iterations=50            &\#iterations=100 \\
    \midrule
    Prior Policy                        & $60.39 \pm 0.9$          & --                        & --                        \\
    PUCT (AlphaGo)                      & $91.92 \pm 0.5$          & $94.62 \pm 0.4$          & $95.25 \pm 0.4$          \\
    PGS (with entropy)                  & $91.26 \pm 0.5$          & $94.85 \pm 0.4$          & $95.58 \pm 0.4$          \\
    \textbf{ExPoSe}                              & $\textbf{94.92} \pm 0.4$ & $\textbf{97.16} \pm 0.3$ & $\textbf{97.39} \pm 0.3$ \\
    \bottomrule
    \end{tabular}
\end{table*}

%% file: tables/hamiltonian.tex
\begin{table*}[ht!]
    \caption{Comparison of test performance on Hamiltonian cycle search (nodes = 50, sparsity = 10\%) measured in success rate (i.e. \% of test instances solved) for different search methods ["\#iterations" stands for number of search iterations]}
    \label{table:hamiltonian-cycle}
    \centering
    \begin{tabular}{c@{\hspace{8\tabcolsep}}c@{\hspace{10\tabcolsep}}c@{\hspace{10\tabcolsep}}c}
    \toprule
    Search Method                       &\#iterations=10           &\#iterations=50            &\#iterations=100 \\
    \midrule
    Prior Policy                        & $01.31 \pm 0.2$          & --                        & --                        \\
    PUCT (AlphaGo)                      & $68.76 \pm 0.7$          & $79.67 \pm 0.6$          & $83.53 \pm 0.6$          \\
    PGS (with entropy)                  & $70.62 \pm 0.7$          & $91.47 \pm 0.4$          & $95.27 \pm 0.3$          \\
    \textbf{ExPoSe}                              & $\textbf{71.27} \pm 0.7$ & $\textbf{93.40} \pm 0.4$ & $\textbf{97.09} \pm 0.3$ \\
    \bottomrule
    \end{tabular}
\end{table*}

%% file: tables/navigation.tex
\begin{table*}
    \caption{Comparison of test performance on grid navigation (hard set) measured in success rate (i.e. \% of test instances solved) for different methods ["\#iterations" stands for number of search iterations]}
    \label{table:navigation}
    \centering
    \begin{tabular}{c@{\hspace{10\tabcolsep}}c@{\hspace{10\tabcolsep}}c@{\hspace{10\tabcolsep}}c}
    \toprule
    Search Method                       &\#iterations=10           &\#iterations=50            &\#iterations=100 \\
    \midrule
    Prior Policy                        & $81.18 \pm 0.6$          & --                        & --                        \\
    PUCT (AlphaGo)                      & $92.38 \pm 0.4$          & $93.44 \pm 0.4$          & $94.02 \pm 0.4$          \\
    % GPUCT                               & $93.20 \pm 0.4$          & $96.67 \pm 0.3$          & $97.60 \pm 0.2$          \\
    PGS (with entropy)                  & $94.67 \pm 0.3$          & $97.78 \pm 0.2$          & $98.91 \pm 0.2$          \\
    \textbf{ExPoSe}                              & $\textbf{99.38} \pm 0.1$ & $\textbf{99.93} \pm 0.1$ & $\textbf{99.96} \pm 0.1$ \\
    \bottomrule
    \end{tabular}
\end{table*}

%% file: tables/atari.tex
\begin{table*}[ht!]
    \caption{Comparison of test performance on a set of 50 Atari game measured in Human Normalised Score (HNS) and Baseline Normalised Score (BNS) averaged over all the games. We use prior policy as baseline to compute Baseline Normalised Score.}
    \label{table:atari}
    \centering
    \begin{tabular}{c@{\hspace{10\tabcolsep}}c@{\hspace{10\tabcolsep}}c}
    \toprule
    Search Method       & Human Normalised Score    & Baseline Normalised Score \\       
                        & Mean(Median)              & Mean(Median)              \\
    \midrule
    Prior Policy        & 3.32 (0.63)               & 1.00 (1.00)               \\
    PGS (with entropy)  & 6.79 (0.86)               & 1.22 (1.27)               \\
    ExPoSe              & \textbf{7.15 (0.96)}      & \textbf{1.32 (1.38)}      \\
    \bottomrule
    \end{tabular}
\end{table*}

%% file: tables/ablation.tex
\begin{table*}
    \caption{Comparison of test performance on Sokoban for the ablation study to analyse the contribution of each improvement described in ExPoSe. We modify the original ExPoSe implementation to create other methods for comparison ["\#iterations" stands for number of search iterations].}
    \label{table:ablation}
    \centering
    \begin{tabular}{c@{\hspace{10\tabcolsep}}c@{\hspace{10\tabcolsep}}c@{\hspace{10\tabcolsep}}c}
    \toprule
    Search Method                                   &\#iterations=10    &\#iterations=50    &\#iterations=100       \\
    \midrule
    \textbf{ExPoSe}                                          & $94.92 \pm 0.4$  & $97.16 \pm 0.3$  & $97.39 \pm 0.3$      \\
    ExPoSe \textit{without Importance Sampling}     & $94.00 \pm 0.4$  & $96.30 \pm 0.3$  & $97.16 \pm 0.3$      \\
    ExPoSe \textit{without Value Baseline}          & $93.14 \pm 0.5$  & $95.61 \pm 0.4$  & $96.41 \pm 0.3$      \\
    ExPoSe \textit{with Value Network Baseline}     & $94.36 \pm 0.4$  & $96.41 \pm 0.3$  & $96.70 \pm 0.3$		\\
    \bottomrule
    \end{tabular}
\end{table*}

%% file: text/related.tex
\section{Related Works}
Decision-making problems with smaller state spaces can be solved exactly using the value iteration algorithm \cite{sutton2018reinforcement}. However, applying value iteration on problems with massive search space, such as chess and go, is intractable. Instead, we can use a parameterised policy or value function for these larger decision-making problems. These parameterised models can be learnt using reinforcement learning algorithms like Q-learning or policy gradients \cite{sutton2018reinforcement}. More recent works have tried to take advantage of neural networks as function approximators \cite{Mnih2013PlayingAW,hessel2018rainbow,schulman2015trust,schulman2017proximal,mnih2016asynchronous} and have reported great success on problems like atari, OpenAI gym~\cite{brockman2016openai}, or Mujoco simulator~\cite{MuJoCo}.

Alternatively, we can use an online search to select an action for the agent's current state. Methods like UCT~\cite{kocsis2006bandit} iteratively expand a search tree using a black-box simulator and output an action. These methods can massively improve their performance when coupled with well-defined heuristics \cite{coulom2007computing}. The importance of exploration in decision making problems has also been studied for both reinforcement learning algorithms \cite{auer2002finite,osband2016deep,fortunato2018noisy,aravindan2021state} as well as online search algorithms \cite{kocsis2006bandit,rosin2011multi}.

Many recent works have tried to combine these two decoupled approaches to leverage their capabilities \cite{tesauro1994td,baxter1998knightcap,silver2016mastering,anthony2017thinking}. For example, AlphaGo~\cite{silver2016mastering} demonstrated that by combining an online tree search method with an offline learnt policy and value function, an agent could beat the world champion in the game of Go, which was seen as a complex challenge for an agent. Following it, AlphaZero~\cite{silver2017mastering} showed that combining online tree search with learnt policy and value function can beat any other agent in chess, shogi and go without prior human knowledge by following a simple training regime defined in \cite{silver2017mastering}. 

Within this research direction, some proposed works focus on learning the environment model and using it for simulation \cite{racaniere2017imagination,silver2017predictron}. For example, Predictron~\cite{silver2017predictron} learns an abstract environment model to simulate a trajectory and accumulate internal rewards. On the other hand, Value Iteration Network~\cite{tamar2016value} and Gated Path Planning Network~\cite{lee2018gated} learn an environment model in the context of a planning algorithm by embedding the algorithmic structure of the value iteration algorithm in the neural network architecture. Furthermore, ATreeQN~\cite{farquhar2018treeqn} learns an environment model in the context of a tree structure and uses it to predict the Q-values. These algorithms jointly optimise the environment dynamics with the policy or value function approximator. Further, Guez et al.~\cite{guez2019investigation} argue that a recurrent neural network could exhibit the properties of a planning algorithm without specifying an algorithmic structure in the network architecture.

Alternatively, MCTSnets~\cite{guez2018learning} tries to learn a parameterised policy in the context of a planning algorithm. It mimics the structure of the MCTS algorithm and learns to guide the search using parameterised memory embeddings stored in a tree structure. Similarly, Pascanu et al.~\cite{pascanu2017learning} also learn to plan using a neural network model, but it uses an unstructured memory representation.
MuZero~\cite{schrittwieser2020mastering} learns the environment model and uses it with an online tree search algorithm like UCT.

In an alternative research direction, some online search methods try to adapt a parameterised policy or value function learnt offline on a large dataset by simulating trajectories from the input state and using model-free RL methods to optimise the parameters of the policy or the value function \cite{silver2009reinforcement,bello2016neural,anthony2019policy}. For example, Policy Gradient Search~\cite{anthony2019policy} follows a Monte-Carlo Search framework while iteratively adapting the simulation policy using policy gradients. Further, entropy regularisation can be used while computing the gradients to prevent the issues of early commitment and initialisation bias \cite{larma2021improving}. Policy gradient methods can also be used to improve the rollout policy, which can further improve the performance of Monte-Carlo Tree Search \cite{graf2015adaptive}.

%% file: text/conclusion.tex
\section{Conclusion}
In this paper, we identify and empirically analyse the key reasons behind the success of tree-based and gradient-based online search methods. Firstly, we find that information sharing across the state space during the online search helps in improving the agent's performance, and Policy Gradient Search provides a practical algorithm to achieve this by directly updating the parameters of the search policy. Secondly, we determine that an explicit exploration mechanism is essential for efficiently balancing the exploration-exploitation trade-off and enabling the online search to escape from a local optimum. While tree-based methods have a well-defined exploration mechanism by design, Policy Gradient Search relies on entropy regularisation to induce exploration by increasing randomness in the policy.

To address these issues, we propose an efficient and effective online search method called Exploratory Policy Gradient Search (ExPoSe), which combines gradient-based policy improvement with state-based exploration. ExPoSe iteratively simulates trajectories using a simulation policy that incorporates an exploration term depending on the state-action visitation count, and updates the parameters of the prior policy using REINFORCE. Furthermore, we discuss the issue of optimising the prior policy using the trajectories generated by the exploration-induced simulation policy naively, which could limit the agent's performance gain due to the off-policy data generation. ExPoSe resolves this problem by using importance sampling to obtain an unbiased estimate of the policy gradients. Additionally, we highlight the benefits of using a tree structure to maintain the value estimates of the states encountered during the online search, which can further reduce the variance of the policy gradients.

We conduct experiments on a diverse set of decision-making problems, including goal-based planning problems like Sokoban, Hamiltonian cycle search in sparse graphs and grid navigation, and image-based games such as Atari 2600. Our experimental results demonstrate that ExPoSe outperforms other online search methods consistently across all test domains. However, ExPoSe, like many other search algorithms, requires access to the environment simulator for the search, which may be infeasible for some problems. Integrating a learned world model into the algorithm could be a potential solution to this limitation, and future work could explore its effectiveness in improving the applicability and robustness of ExPoSe.

\section*{Acknowledgement}
This research is supported by the National Research Foundation Singapore and DSO National Laboratories under the AI Singapore Programme (Award Number: AISG2-RP-2020-016).

We would like to acknowledge the usage of ChatGPT for helping in correcting the grammar in this paper.